\documentclass[runningheads]{llncs}
\usepackage{graphicx}
\usepackage{comment}
\usepackage{amsmath,amssymb} 
\usepackage{color}

\usepackage{hyperref}
\usepackage{subcaption}
\usepackage{setspace}
\usepackage{amssymb}

\begin{document}
\pagestyle{headings}
\mainmatter
\def\ECCVSubNumber{918}  

\title{Facing the Hard Problems in FGVC} 

%
\author{Connor Anderson \and
Matt Gwilliam \and
Adam Teuscher \and
Andrew Merrill \and
Ryan Farrell}
\authorrunning{C. Anderson et al.}
%
\institute{Brigham Young University, Provo UT 84602, USA\\
\email{connor.anderson@byu.edu}\;
\email{farrell@cs.byu.edu}\;
\email{\{mattgwilliamjr,adam.m.teuscher,andrewmerrill215\}@gmail.com}\;
}
\maketitle

\begin{abstract}

In fine-grained visual categorization (FGVC), there is a near-singular focus in pursuit of attaining state-of-the-art (SOTA) accuracy.  This work carefully analyzes the performance of recent SOTA methods, quantitatively, but more importantly, qualitatively.  We show that these models universally struggle with certain ``hard'' images, while also making complementary mistakes.  We underscore the importance of such analysis, and demonstrate that combining complementary models can improve accuracy on the popular CUB-200 dataset by over 5\%.  
In addition to detailed analysis and characterization of the errors made by these SOTA methods, we provide a clear set of recommended directions for future FGVC researchers.

\end{abstract}

%
%

\vspace{-.55in}

\begin{figure}[h]
    \centering
    \includegraphics[width=\linewidth]{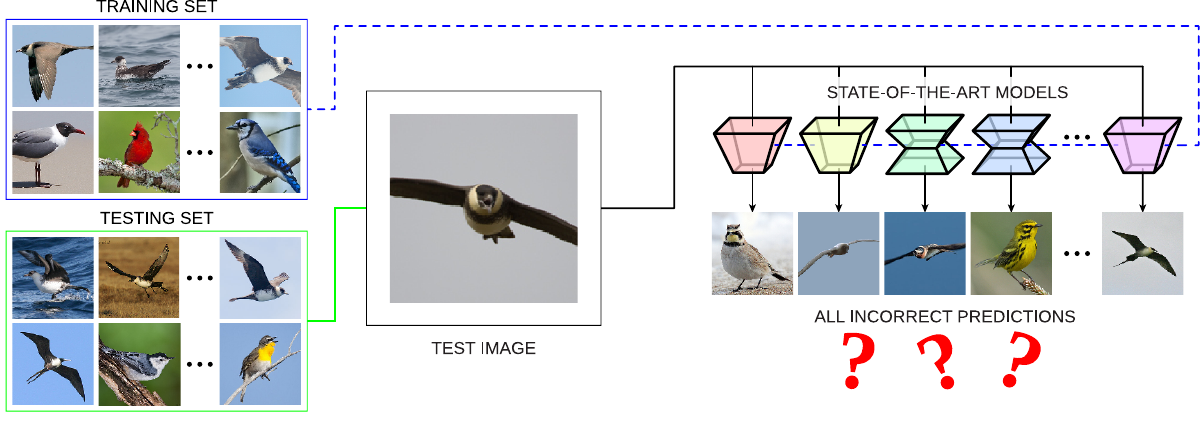}
    \caption{\small\textbf{Motivation - All State-of-the-art Models Incorrect!} Despite cutting-edge performance, there are images that  state-of-the-art models universally predict incorrectly.  This is a real example where an image of a Pomarine Jaeger (from the CUB dataset) is incorrectly classified by all state-of-the-art-models (the prediction images are the actual classes predicted).  This is most likely due to a lack of training images for Pomarine Jaeger in that pose.  We assert that cross-pose correspondence, which humans readily perform, is an important direction for future FGVC research.}
    \label{fig:motivation}
\end{figure}

\vspace{-.375in}

\section{Introduction}


Fine-grained visual categorization (FGVC), with its high intra-class variation (e.g. different poses) and low inter-class variation (e.g. all members have the same parts), is a uniquely challenging task within the broader area of object recognition/image classification. While the deep learning approaches developed recently (see discussion in Sec.~\ref{sec:related-work}) have improved benchmark performance considerably, they still have significant issues. Even the best methods, once trained, mislabel images that a human expert would not. When applied to real world data, these models leave much to be desired. In this paper, we seek to study these methods' weaknesses. Rather than looking at model architectures or feature maps, we go straight to the errors themselves, performing rigorous analysis on the per-instance mistakes of state-of-the-art (SOTA) FGVC methods.

Figure~\ref{fig:motivation} provides a great example of one of the key types of mistakes that these networks make. 
The struggle that classifiers have with this image serves as a reminder that their reliance on training data is both a gift and a curse.  While increasing the amount of good data is a reliable way to increase performance, lack of representation of a given pose (such as a bird in flight) or other variation in the training data often guarantees a model will miss when that pose or variation occurs in the test set.

The fact that many such issues exist is well-documented and understood within the community. Our purpose in this work is not to simply restate that these problems exist, but rather to document, both qualitatively and quantitatively, the degree to which a broad set of issues manifest themselves in the data, and how they actually affect the performance of SOTA methods. Our analysis looks beyond the standard measure of overall accuracy, a practice that we believe to be critical for further advancement of the field.

In our effort to directly confront these long-standing issues, this work makes the following contributions:
\begin{itemize}
\item We discuss a framework for carefully analyzing and determining which individual images are the most difficult to classify (see Sec.~\ref{subsec:hardimages}).
\item We perform extensive evaluation of 10 representative FGVC methods (5 baseline and 5 state-of-the-art) across 8 datasets.  This assessment includes quantitative comparison between methods and detailed analysis of the overlap in their prediction errors.
\item Drawing on the overlap analysis, we create simple ensembles of the top methods, achieving categorization accuracies on CUB as high as 95.34\%.  This is quite remarkable as it is nearly 5\% better than the highest published accuracy to-date. \footnote{Krause \textit{et al.}~\cite{KrauseSHZTDPF_ECCV2016} report 92.8\%, however, they leverage an additional large-scale dataset consisting of millions of images.}  We're not claiming a new method with SOTA performance, rather we are highlighting how effective leveraging the complementary analysis can be.
\item We define a grouping for fine-grained visual categorization errors.  Using a novel real-world birds dataset, iCUB (images of CUB categories gathered from iNaturalist.org), together with CUB++, a cleaned version of CUB, we present both qualitative and quantitative analysis of the mistakes that SOTA methods make.
\end{itemize}
Perhaps most importantly -- in addition to closely analyzing individual errors -- we take a bird's eye view, converging on a clear set of recommendations for future researchers in the FGVC community.


%
%

\section{Related Work}\label{sec:related-work}

For general image classification, categories can be distinguished based on major differences such as the presence or absence of key parts (e.g. a human has legs while a car has wheels). In FGVC, however, categories are primarily differentiated by subtle differences in the shape or color of parts that they share in common (e.g. differences in beak color or length between different sea bird species). While large datasets like ImageNet~\cite{DengDSLLF_CVPR2009,RussakovskyDSKSMHKKBBF_IJCV2015} and iNat~\cite{van2018inaturalist} have subdomains that are fine-grained, FGVC research more commonly uses single-domain datasets such as Aircraft~\cite{MajiRKBV_arXiv2013}, Cars~\cite{KrauseSDF_ICCVW2013}, Dogs~\cite{KhoslaJYF_CVPRW2011}, Caltech Birds~\cite{WahBWPB_Tech2011}, Flowers~\cite{NilsbackZ_CVPR2006}, and North American Birds~\cite{VanHornBFHBIPB_CVPR2015} as benchmarks, which each have hundreds of classes and dozens of images per class.

Deep learning has become the dominant approach to FGVC~\cite{Chen_2019_CVPR,DubeyGRN_NeurIPS2018,Ge_2019_CVPR,Luo_2019_ICCV,WangMD_CVPR2018,ZhengFML_ICCV2017}, like computer vision in general. Most models are pre-trained on ImageNet~\cite{DengDSLLF_CVPR2009,RussakovskyDSKSMHKKBBF_IJCV2015}, while some approaches train on other, related datasets before the target FGVC task (e.g.~\cite{KrauseSHZTDPF_ECCV2016}). Data augmentation has become ubiquitous in fine-grained algorithms, with recent work exploring how to optimize what transformations are used~\cite{Cubuk_2019_CVPR}.

Recently, several main approaches to FGVC have emerged. Segmentation \cite{ChaiLZ_ICCV2013,yao2016coarse}, part-model~\cite{huang2016part,SimonR_ICCV2015}, and pose-alignment~\cite{GavvesFSST_ICCV2013,GavvesFSST_IJCV2015,GoringRFD_CVPR2014,GuoF_WACV2019,LiuZNSHLFYHM_ECCV2018} methods attempt to isolate and model important class-specific features in a pose-invariant way. Recent pooling methods -- such as bilinear pooling~\cite{KongF_CVPR2017,LinRM_PAMI2017,LinM_BMVC2017,YuZZZY_ECCV2018} and the related Grassman pooling~\cite{WeiGZZZ_ECCV2018}, covariance pooling~\cite{LiXWG_CVPR2018,li2017second}, and several learnable pooling~\cite{CuiZWLLB_CVPR2017,SimonGDDR_ICCV2017} methods -- attempt to leverage second order statistics between deep CNN features. Attention methods~\cite{RodriguezGCRG_ECCV2018,WangJQYLZWT_CVPR2017} have also received a lot of ``attention'', with different methods employing recurrent models~\cite{FuZM_CVPR2017,SermanetFR_ICLRW2015,ZhaoWFPY_TMM2017}, reinforcement learning~\cite{LiuWWDL_AAAI2017}, metric learning~\cite{SunYZD_ECCV2018}, and part discovery~\cite{HuQHL_arXiv2019,ZhengFML_ICCV2017,ZhengFZL_CVPR2019}. Other approaches modify the learning objective to account for high similarity and ambiguity between classes. Label smoothing~\cite{SzegedyVISW_CVPR2016,muller2019does} and taxonomy-based schemes~\cite{bertinetto2019making,trammell2019contextual} redistribute probability mass in the target distributions. Pairwise confusion~\cite{DubeyGGRFN_ECCV2018} and maximum entropy~\cite{DubeyGRN_NeurIPS2018} help reduce overconfidence by regularizing predictions.



%
%

\section{Analysis Framework}\label{sec:approach}

In this section, we discuss our framework for identifying the most challenging images in a dataset. Our framework consists of training multiple models on the same dataset and measuring the overlap in their predictions, which partitions the data into sets of varying difficulty. We also discuss our ensembling method, which we use to show that while different state-of-the-art methods are complementary, they still collectively fail on the most challenging subset of images.

\subsection{Discovering Hard Images through Prediction Overlap}
\label{subsec:hardimages}

To discover difficult-to-classify images, we compare the predictions of current SOTA methods. By looking at the overlap in predictions -- and particularly the set of images that none of the models can classify correctly -- we can get a qualitative understanding of where current methods struggle in general.

To do this, we train models using multiple SOTA methods and save their test set  predictions -- the pre-softmax class-score vectors for each image. Using these predictions, we divide the test set images into several groups, based on how many of the models predict them correctly. This places each image on a spectrum of difficulty. Images that all models classify correctly are easy to learn, while images that none of the models classify correctly are inherently challenging. 

To be precise, we assign each image $\mathbf{x_i}$ an overlap label $o_i \in \{0,1,\dots,N\}$, where $N$ is the number of models under consideration (we use five in our experiments): $o_i=\sum_{k=1}^N \mathbf{1}(y_i, \hat{y}_i^k)$. $y_i$ is the ground-truth class label, $\hat{y}_i^k$ is the predicted label from the $k$th model, and $\mathbf{1}(a,b)=1$ if $a=b$ else $0$.

We measure prediction overlap in two settings: \textit{within-method} and \textit{between-method}. For within-method overlap, we use multiple models trained using the same method. For between-method overlap, we use a single model from several different methods. Sec.~\ref{sec:exper-analysis} presents our prediction overlap results, as well as qualitative analysis on which images are challenging and why.

\subsection{Ensemble Methods}\label{subsec:ensemble-methods}


A common method for improving classification performance is to ensemble multiple models, making decisions based on aggregating their individual predictions. However, if none of the individual models is able to make the correct predictions for a given set of images, then combining them will also fail. Our prediction-overlap analysis shows that there are sets of images which none of the SOTA models are able to predict correctly, thus placing a cap on ensemble performance. But the same analysis also shows that there are other subsets of images which are sometimes predicted correctly, but not always for all models. We also find that the set of hard images is smaller in the case of between-method overlap. This leads us to conclude that there are still gains to be had using ensembles, and that combining models from different methods will yield the best results.

To validate our conclusions, we create ensembles -- both within-method and between-method. We use two simple techniques. The first is the \textit{majority-vote ensemble}: the highest-confidence prediction from each model is used to vote for the image class, and the class with the most votes is chosen. We can write this as $\hat{y}_i = \arg\max\left[\text{bincount}(\hat{y}_i^1, \hat{y}_i^2,\dots,\hat{y}_i^N)\right]$, where bincount returns an array of length $C$ (number of classes) indicating the number of models that predicted each class for image $\mathbf{x}_i$. The second is the \textit{averaged-probability ensemble}: the class-probability vectors for an image are averaged across models, and the highest probability class is chosen. In this case, $\hat{y}_i = \arg\max\left[\frac{1}{N}\sum_{k=1}^{N}\mathbf{s}_i^k\right]$, where $\mathbf{s}_i^k$ is the softmax probabilities for image $\mathbf{x}_i$ produced by model $k$. These ensembles provide clear performance improvements over the individual models, despite their simplicity (they require no additional training). We stress, however, that their performance is fundamentally limited by the set of hard samples.

\section{Baseline and State-of-the-art FGVC Methods}\label{sec:methods}

To facilitate thorough evaluation and analysis (see Sec.~\ref{sec:exper-analysis}), we consider a variety of FGVC methods, both baseline and state-of-the-art approaches.


\subsection{Baseline Methods}

We evaluate the following baseline models on each of the fine-grained datasets (see Sec.~\ref{sec:datasets}). Some of these baselines are quite strong -- we call them baselines only to indicate that they are generic image classification models, without any FGVC-specific additions to the models or training procedure. Each model is fine-tuned from ImageNet~\cite{DengDSLLF_CVPR2009}-pretrained weights. We use models and weights publicly available through the PyTorch \texttt{torchvision} module, except where otherwise noted.

\paragraph{\textbf{Res50}} We use the ResNet-50 model originally proposed in~\cite{HeZRS_CVPR2016} -- it is one of the most common baseline architectures in both FGVC and general image recognition.

\paragraph{\textbf{Res50+}} We slightly adapt the basic ResNet-50 model by swapping the final global average pooling layer for global max pooling, and add a batch norm~\cite{ioffe2015batch} layer before the linear classifier. This minor change leads to faster convergence and better final accuracy.

\paragraph{\textbf{ResNeXt}} We use the ResNeXt-50-32x4d described in~\cite{xie2017aggregated}. ResNeXt extends the ResNet architecture with a cardinality dimension -- the residual layers are divided into groups which each operate on a subset of the input features.

\paragraph{\textbf{DenseNet}} We use the DenseNet-161 model described in~\cite{HuangLW_CVPR2017_DenseNet}. DenseNet replaces the additive residual transforms of ResNet with feature concatenation -- every layer in the network takes as input the feature outputs of all preceding layers.

\paragraph{\textbf{ENet}} We use the EfficientNet-b4 model described in~\cite{tan2019efficientnet}. EfficientNets are a family of models developed through neural architecture search to be more accurate and efficient than other architectures of the same size. We use a public PyTorch implementation\footnote{\url{https://github.com/lukemelas/EfficientNet-PyTorch}}, which includes ImageNet-pretrained weights.

\subsection{State-of-the-art Methods}

In addition to the baseline models described above, we analyze the performance of several recent state-of-the-art FGVC models. There are many such models in the literature. We selected a subset of recent work that provides good coverage of the main approaches to FGVC (see Sec.~\ref{sec:related-work}) and for which there were publicly available PyTorch models. We use the authors original implementation where available. For each of these methods we use a ResNet-50 backbone unless otherwise noted.


\vspace{-2.5mm}
\paragraph{\textbf{WS-DAN}} The Weakly Supervised Data Augmentation Network~\cite{HuQHL_arXiv2019} uses attention for data augmentation. The augmentation occurs in two ways: first, attention is used to crop and enlarge salient parts of the image; second, attention is used to mask out salient parts of the image, forcing the network to diversify the features it relies on. At test time, predictions on the full image and attention-crops are fused together.

\vspace{-2.5mm}
\paragraph{\textbf{S3N}} Selective Sparse Sampling~\cite{DingZZYJ_ICCV2019} is another attention-based method. The class peak response map for an image is used to dynamically sample sparse sets of discriminative and complementary image regions, which are then resampled for better focus by the network.

\vspace{-2.5mm}
\paragraph{\textbf{MPN-COV}} Matrix Power Normalized Covariance~\cite{li2017second} is a second-order pooling method. Instead of representing each feature in the final convolution layer of a network as a single value, MPN-COV calculates the covariance relationship between features, which is used for classification. A fast iterative method was proposed in~\cite{LiXWG_CVPR2018}, which we use in our experiments.

\vspace{-2.5mm}
\paragraph{\textbf{DCL}} Destruction and Construction Learning~\cite{ChenBZM_CVPR2019} is a multi-objective method to help the classifier learn discriminative contextual features. The images are divided into regions and then scrambled before being processed by the network. One network branch attempts to classify the scrambled image, while another attempts to unscramble it. An additional adversarial branch tries to distinguish between regular and scrambled images.

\vspace{-2.5mm}
\paragraph{\textbf{MaxEnt}} The Maximum Entropy~\cite{DubeyGRN_NeurIPS2018} loss function is used to improve generalization by reducing prediction over-confidence. Maximum entropy extends the regular cross-entropy loss by adding a penalty for low-entropy prediction distributions. We use the maximum entropy loss function in conjunction with a DenseNet-161 model, the top performing configuration reported in~\cite{DubeyGRN_NeurIPS2018}.





\section{Datasets} \label{sec:datasets}

\begin{table}[b]
    \centering
    {\small
     \setlength{\tabcolsep}{1.75mm}
    \begin{tabular}{p{45mm}p{20mm}ccc} 
         \hline
         \textbf{Dataset} & & \textbf{Classes} & \textbf{\#Train} & \textbf{\#Test} \\ 
         \hline
         FGVC Aircraft & (Aircraft) & 100 & 6,667 & 3,333 \\
         Stanford Cars & (Cars) & 196 & 8,144 & 8,041 \\
         Caltech-UCSD Birds-200-2011 & (CUB) & 200 & 5,994 & 5,794 \\
         Stanford Dogs & (Dogs) & 120 & 12,000 & 8,580 \\
         Oxford Flowers 102 & (Flowers) & 102 & 2,040 & 6,149 \\
         Birds of North America & (NABirds) & 555 & 23,929 & 24,633 \\
        \hline
         CUB++ & & 200 & 5,875 & 5,661 \\
         iCub100 & & 200 & n/a & 19,698 \\
         \hline
    \end{tabular}
    }
    \caption{\small Number of classes, training images, and test images for each of the eight datasets considered in this paper. Aircraft, Cars, CUB, Dogs, Flowers, and NABirds are popular, established FGVC datasets. We introduce the CUB++ and iCub100 datasets to aid in our analysis -- see the text for details.}
\label{table:datasets}
\end{table}

\begin{figure}[t]
    \centering
    \includegraphics[width=\linewidth,height=.5\linewidth]{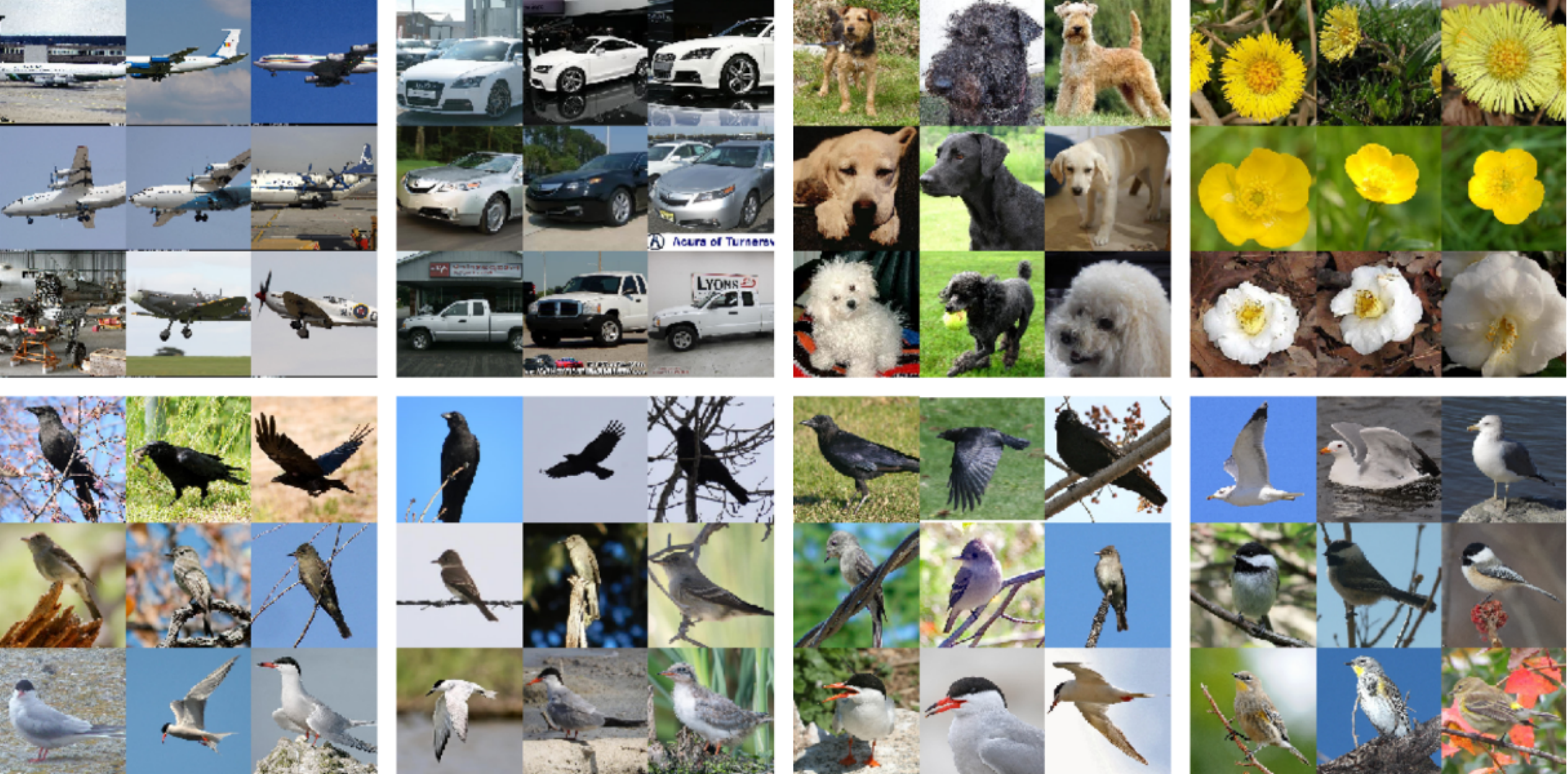}
    \caption{\small\textbf{Datasets - Example Images.} From left-to-right, top-to-bottom, images from 3 classes each of: Aircraft, Cars, Dogs, Flowers, CUB, iCUB, and two sections for NABirds (so 6 classes). The first 3 sections on the bottom contain the same 3 species.}
    \label{fig:datasets}
\end{figure}

We perform analysis on eight fine-grained datasets, using the architectures and methods described in Sec.~\ref{sec:methods}. We use six established FGVC datasets -- Aircraft~\cite{MajiRKBV_arXiv2013}, Cars~\cite{KrauseSDF_ICCVW2013}, CUB~\cite{WahBWPB_Tech2011}, Dogs~\cite{KhoslaJYF_CVPRW2011}, Flowers~\cite{NilsbackZ_ICVGIP2008,NilsbackZ_IVC2010}, and NABirds~\cite{VanHornBFHBIPB_CVPR2015} -- as well as two additional datasets that we created to aid in our analysis. We call these two additional datasets CUB++ and iCub100 (described below). Table~\ref{table:datasets} gives information on each of these datasets, and example images are shown in Fig.~\ref{fig:datasets}.

\begin{figure}[t]
    \centering
    \includegraphics[width=\linewidth]{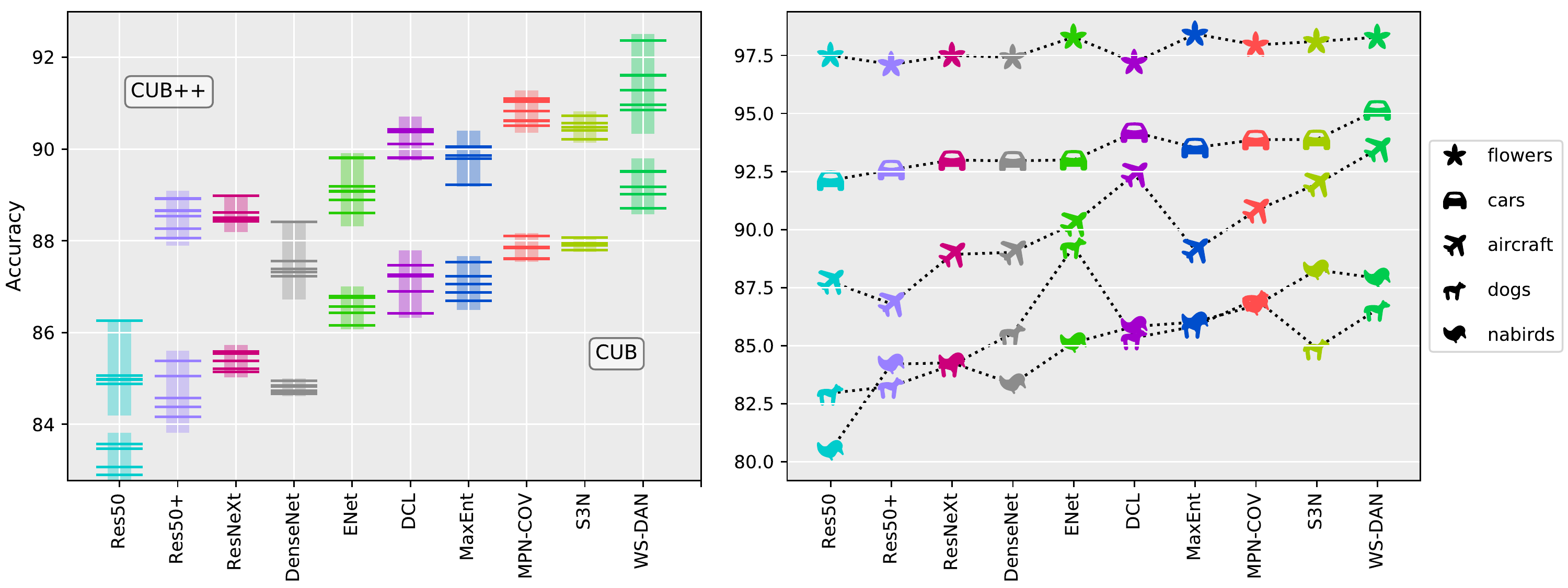}
    \caption{\small\textbf{Summary of model accuracy}. On the left we show five models for each of the methods in Sec.~\ref{sec:methods} (horizontal ticks), along with 2 standard deviations around the mean (vertical bars) on CUB and CUB++. On the right we show performance of a single model for each method on five other FGVC datasets.}
    \label{fig:tickplots}
\end{figure}

In ~\cite{VanHornBFHBIPB_CVPR2015}, Van Horn et al. discovered an error rate of about 4\% in the labels of CUB. They also obtained the corrected labels, by the same process they used to annotate NABirds. In order to remove the noise from our analysis as much as possible, we obtained the corrected labels from the authors of~\cite{VanHornBFHBIPB_CVPR2015} and used them to create CUB++. CUB++ uses the same images as CUB, but the erroneous labels have been corrected. In addition, we removed images for which the corrected label was not one of the original CUB labels; there were 119 such images in the training set, and 133 in the test set. Our experimental results show that using the fixed set of data provides a non-trivial jump in performance of 2-3\% for all methods considered (see Fig.~\ref{fig:tickplots}).

In addition to CUB++, we collected a new dataset which we call iCub100. The dataset consists of up to 100 images from each of the 200 categories in CUB (several of the categories have fewer than 100 images available). The images are a random sample of the research-grade images for these categories on the iNaturalist website. We treat iCub100 as an additional large-scale validation set, and use it to study model generalization and failure modes. We don't use any of the iCub100 data for training. The iCub100 data is real-world and ``raw": we didn't filter the data. This means that some of the images are low quality or may be missing the target subject: for instance, some images are of nests and eggs. However, most of the images contain recognizable subjects from the appropriate class.

\section{Experiments and Analysis} \label{sec:exper-analysis}

Using the methods and datasets discussed in Sec.~\ref{sec:methods} and~\ref{sec:datasets}, we perform a detailed analysis with the goal of discovering where current SOTA methods fail in terms of the images they struggle to classify correctly. We employ the analysis framework proposed in Sec.~\ref{sec:approach}, and find that each dataset contains a \textit{hard subset} of images, which none of the current SOTA methods are able to correctly classify. In addition, we find that different methods often make different mistakes, so that ensembles of different methods are particularly effective -- but they can't overcome the fundamental limitation of the hard subset. Finally, we perform a qualitative analysis of the hard subset of images in CUB++ and iCub100 and attempt to group these images into several error classes with specific properties.


\subsection{Prediction Overlap}\label{subsec:pred-overlap}

\begin{figure}[t]
    \centering
    \begin{subfigure}{\linewidth}
        \includegraphics[width=\linewidth]{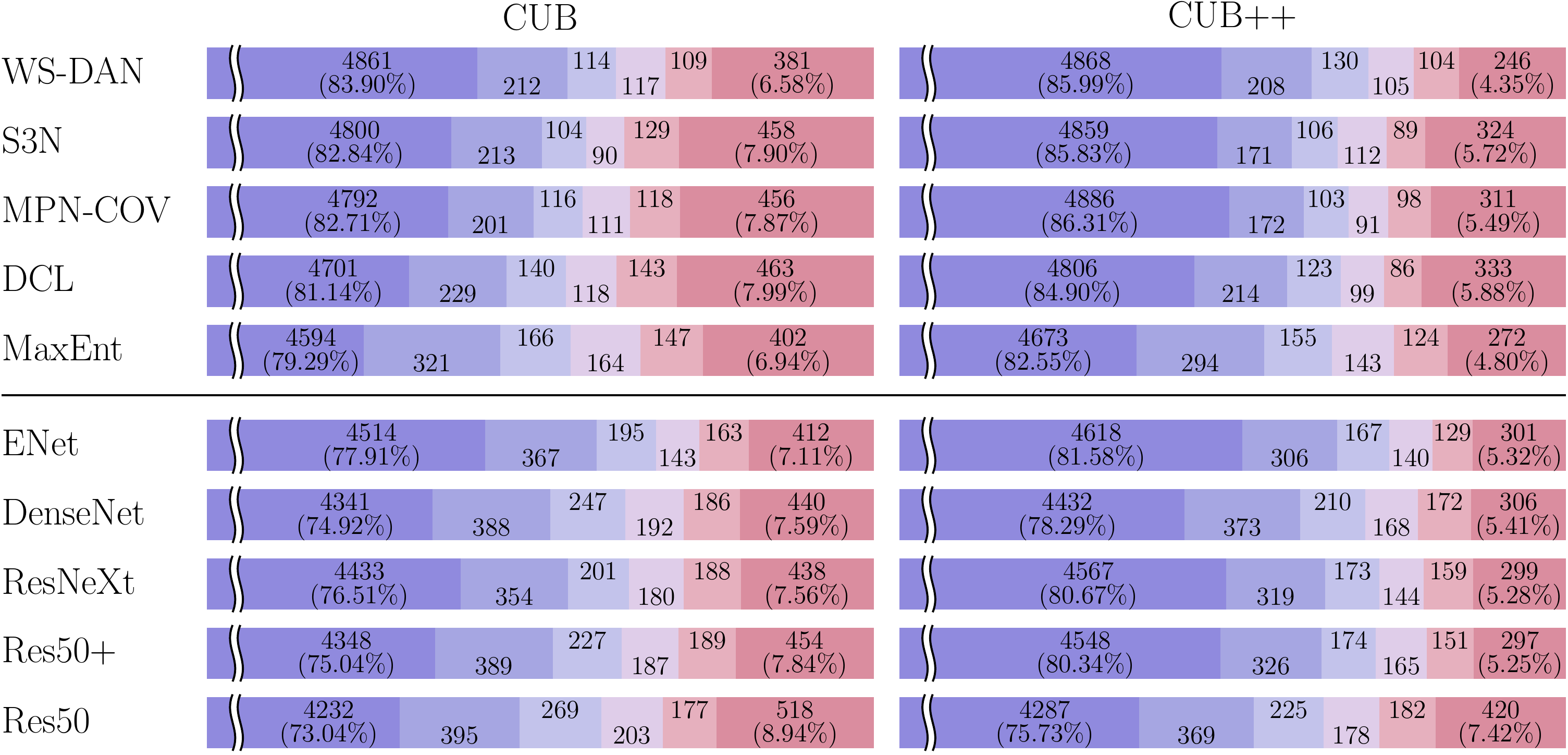}
        \caption{\small{\textbf{Within-method} prediction overlap}
                 \vspace{1mm}}
        \label{fig:within-overlap}
    \end{subfigure}
    \begin{subfigure}{\linewidth}
        \includegraphics[width=\linewidth]{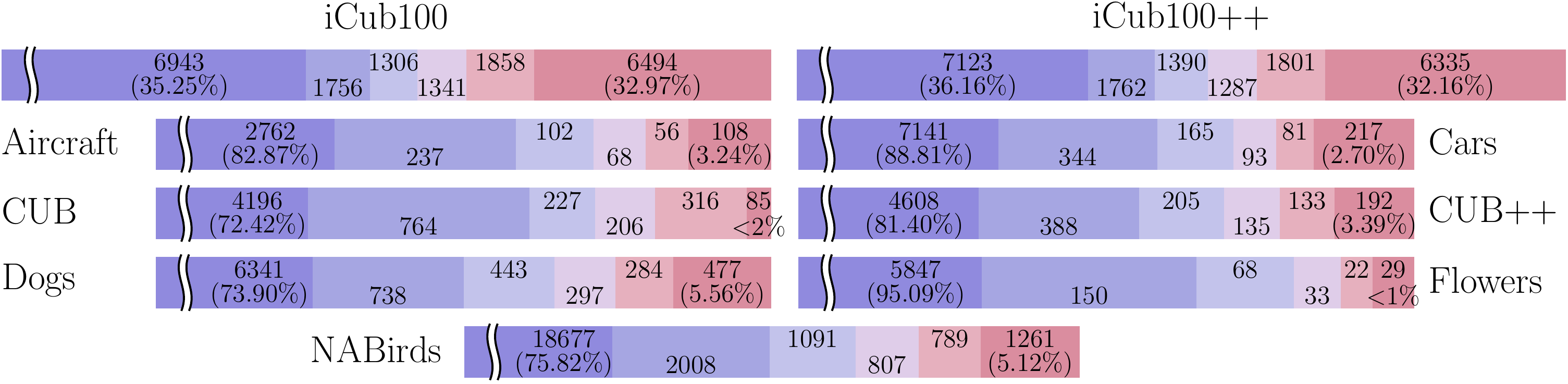}
        \caption{\small \textbf{Between-method} prediction overlap}
        \label{fig:between-overlap}
    \end{subfigure}
    \caption{\small\textbf{Prediction overlaps}. (\subref{fig:within-overlap}) Overlap of 5 models from each method on CUB and CUB++. Each bar shows, from left to right, the number of test images correctly predicted by five down to zero models. The methods are ordered by performance on CUB. (\subref{fig:between-overlap}) Overlap of a single model from each SOTA method on each dataset. iCub100 has no training set -- instead, we use models trained on CUB and CUB++ (iCub100 and iCub100++).}
    \label{fig:pred-overlap}
\end{figure}

To obtain prediction overlaps, we train models on each dataset (Sec.~\ref{sec:datasets}) using each of the considered methods (Sec.~\ref{sec:methods}). For comparison, the classification performance of these models are shown in Fig.~\ref{fig:tickplots}. As described in Sec.~\ref{subsec:hardimages}, we measure both within-method and between-method overlap. To measure prediction overlap, we simply count, for each image, the number of models that predicted the correct class, and then group images by this count. The prediction overlaps are summarized in Fig.~\ref{fig:pred-overlap}.


\vspace{-2.5mm}
\paragraph{\textbf{Within-Method Overlap}}
In Fig.~\ref{fig:within-overlap}, we show the within-method prediction overlap using five different models for each method on CUB and CUB++. All models from a given method are trained with the same hyper-parameters: differences in final predictions are due to randomness in training, such as mini-batch sampling and data augmentation. 

We make the following observations about within-method overlap. First, overall performance (accuracy) is strongly correlated with the size of the ``easy'' set of images within a method -- the set of images which all of the models predict correctly. Second, the easy set contains the majority of the images -- from 73\% (Res50, CUB) to 86.3\% (MPN-COV, CUB++) -- which indicates that there is a decent degree of consistency between models within the same method. However, there are a non-trivial number of images in each overlap subset, which brings us to our third observation: across all the tested methods, 6-9\% of images in CUB and 4-7\% of images in CUB++ are \textit{never} classified correctly -- furthermore, the subset of images that are never classified correctly is strictly larger than the subsets that one, two, or three models get correct, and usually larger than the subset that four get correct. This provides our first evidence for \textit{hard subsets} -- for any given method, there are images that are just too difficult to classify correctly.



\vspace{-2.5mm}
\paragraph{\textbf{Between-Method Overlap}}

In Fig.~\ref{fig:between-overlap}, we show the between-method prediction overlap. For each dataset, we train a single model from each of the five SOTA method (WS-DAN, S3N, MPN-COV, DCL, MaxEnt) and measure their prediction overlap. For iCub, which has no training set, we evaluate using a set of models trained on CUB and another trained on CUB++.

Similar to within-method overlap, we see that the majority of the images are classified correctly by all models -- with the exception of iCub, for which overall performance is low (around 50\% accuracy). If we compare the easy subsets of CUB and CUB++ from the between-method overlap to the within-method overlap, however, we see that the easy subsets are in fact \textit{smaller}. In addition, the hard subsets are \textit{also smaller} -- but they don't disappear. From these realizations, we can draw several conclusions. First, \textit{different methods are complementary in their errors}. While each method has an easy and a hard subset, those subsets don't completely overlap across methods. Second, despite their complementary nature, \textit{some images are challenging for all methods} -- that is, there is a hard subset even when you combine multiple complementary models. However, the hard subset is smaller than if you use multiple models from the same method, suggesting that combining models from different methods should yield higher gains than combining models from the same method. We validate this insight in the next section on ensembles.

\begin{table}[t]
    \centering
    {\small
    \setlength{\tabcolsep}{1.5mm}
    \begin{tabular}{p{16mm}p{16mm}ccccc}
    \hline
        \textbf{Dataset} & \textbf{Type} & \textbf{MaxEnt} & \textbf{DCL} & \textbf{MPN-COV} & \textbf{S3N} & \textbf{WS-DAN} \\
         \hline
        \textbf{CUB} & \textit{Single} & 87.080 & 87.056 & 87.856 & 87.929 & 89.189 \\
        & Vote & 88.298 & 87.884 & 88.574 & 88.712 & 89.921 \\
        & cp-Avg & \textbf{88.367} & \textbf{88.091} & \textbf{88.661} & \textbf{88.799} & \textbf{90.007} \\
        \hline
        \textbf{CUB++} & \textit{Single} & 89.793 & 90.228 & 90.821 & 90.479 & 91.418 \\
        & Vote & \textbf{91.168} & 91.203 & 91.486 & 90.973 & 92.369 \\
        & cp-Avg & \textbf{91.168} & \textbf{91.221} & \textbf{91.645} & \textbf{91.397} & \textbf{92.422} \\
        \hline
    \end{tabular}
    }
    \caption{\small \textbf{Within-method ensembles}. Ensemble accuracy is shown for both voting and class-probability-average (cp-Avg) ensembles (see Sec.~\ref{sec:approach}). The average single-model performance is also shown for reference.}
    \label{tab:ensemble-within}
\end{table}

\subsection{Ensembles}

As described in Sec.~\ref{sec:approach}, we create simple ensembles using voting and class-probability averaging. Both of these approaches are applied using the independent model predictions without any need for further training. Our purpose is to see how performance is affected by both the complementary nature of the different methods as well as the existence of the hard samples.



\vspace{-2.5mm}
\paragraph{\textbf{Within-Method Ensembles}} \label{sec:self-ensemble}

Table~\ref{tab:ensemble-within} shows the results of within-method ensembles for CUB and CUB++, created using five models for each method. As predicted by the overlap analysis, these ensembles provide decent performance gains over the average performance of any single model. For example, the average individual model accuracy for WS-DAN on CUB++ is 91.42\% -- ensembling those models gives an accuracy of 92.42\%. We see similar gains of around 1\% for the other methods as well. Voting and probability-averaging yield similar results.

\begin{figure}[t]
    \centering
    \begin{subfigure}[b]{0.48\linewidth}
      \includegraphics[width=\linewidth]{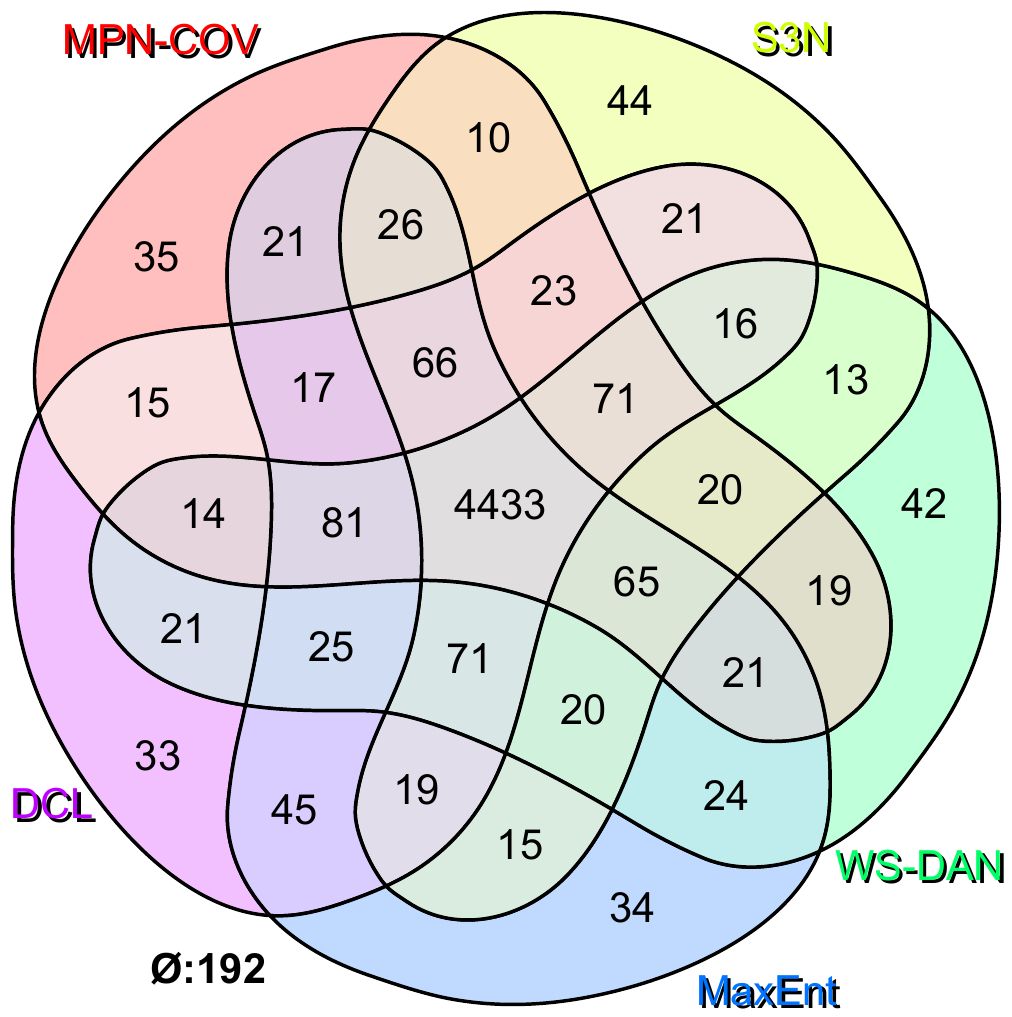}
      \caption{Method Overlaps on CUB++}
      \label{fig:method-overlaps}
    \end{subfigure}
    \begin{subfigure}[b]{0.48\linewidth}
      \includegraphics[width=\linewidth]{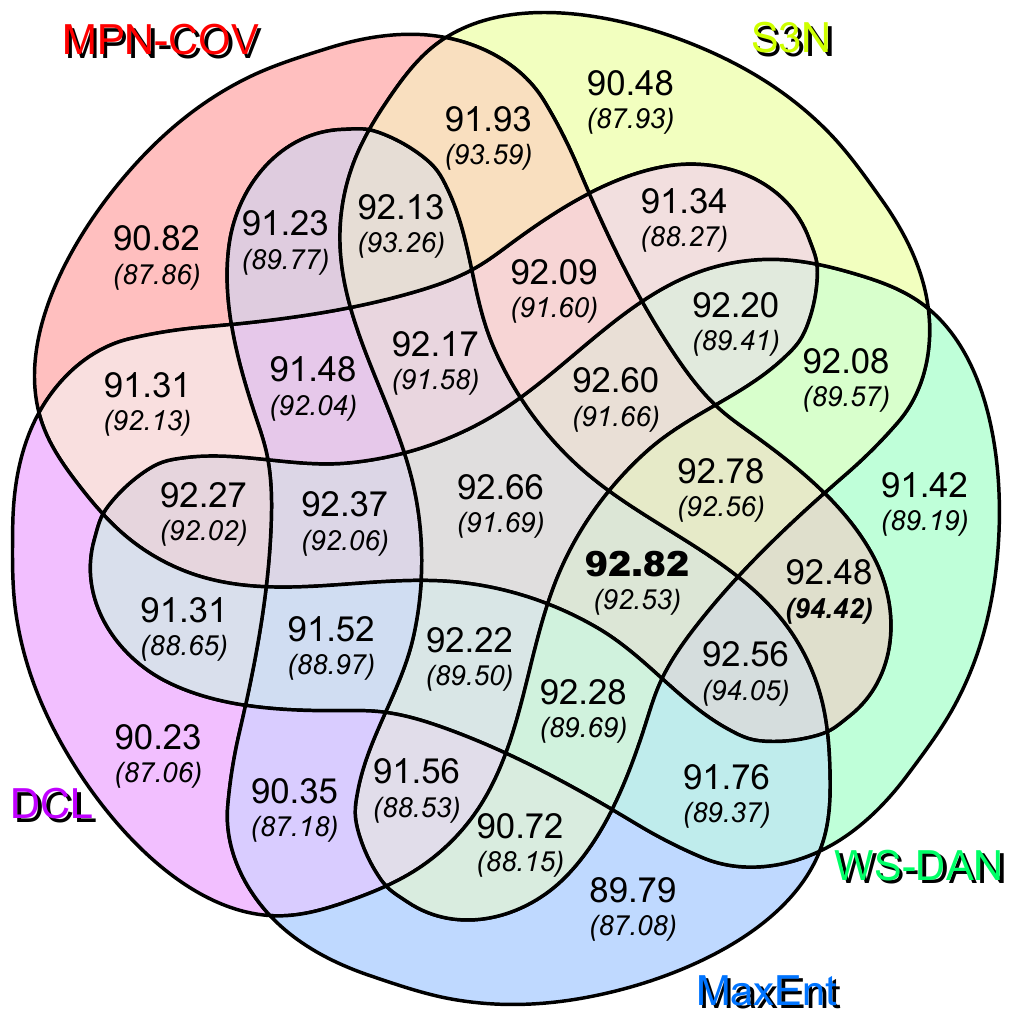}
      \caption{Ensemble Perf. on CUB++/\emph{(CUB)}}
      \label{fig:ensemble-perf}
    \end{subfigure}
    \caption{\small\textbf{Method Overlaps and Ensemble Performance.}
    (\subref{fig:method-overlaps}) Shows the number of images that subsets of the SOTA methods get correct.  Note that there are 192 images that none of these state-of-the-art methods predict correctly ($|\varnothing| = 192$).
    (\subref{fig:ensemble-perf}) Shows the performance of ensemble subsets of the SOTA methods using class-probability averaging.  Note that there are two numbers in each cell: the top one is the performance on CUB++; the lower one, parenthesized in italics, is the performance on CUB.}
    \label{fig:overlaps_methods}
\end{figure}

\vspace{-2.5mm}
\paragraph{\textbf{Between-Method Ensembles}} \label{sec:sota-ensemble}

Fig.~\ref{fig:ensemble-perf} shows the between-method ensemble performance on CUB and CUB++ when taking a single model from each subset of the SOTA methods. These results are averaged over five separate and disjoint ensembles, using five different models from each method (twenty-five total models). Again, as predicted by the overlap analysis, using models from different methods results in better performance than using models from the same method. Furthermore, by using five models each from WS-DAN and MPN-COV, \textit{we achieve a remarkable 95.53\% accuracy on CUB}, using a probability-average ensemble. A single WS-DAN and MPN-COV model together still provide excellent results, at 94.42\%.

There are a few things we would like to point out from Fig.~\ref{fig:ensemble-perf}. First, MPN-COV seems to be the most ``complementary'' model that we tested -- ensembles including MPN-COV consistently perform better than those without. Second, and curiously, ensembles on CUB tend to outperform those on CUB++ -- sometimes by a large margin. This was unexpected, as CUB++ performs much better than CUB at the single-model level, as well in within-method ensembles. As the only difference between CUB and CUB++ is the removal of mislabeled images, we are left to conclude that something about the \textit{noise} in the CUB labels leads to more complementary models. This is a question that merits further exploration.

Finally, if we compare the ensemble predictions to the between-method overlap groups, we find that the ensembles correctly predict most of the images in the overlap groups of between three and five models, but get far fewer of the images in the overlap-two group, and almost none of the overlap-one and zero groups. This further supports the main claim of the paper: that there are inherently challenging subsets of images that current methods are incapable of learning.


\subsection{Qualitative Analysis} \label{subsec:qualitative}

We now turn our focus to a qualitative analysis of the hard-image subsets. We use birds as a case study, due to their prevalence in FGVC literature. In particular, we use the between-method hard subsets of CUB++ and iCub100. We manually inspect the challenging images, and attempt to group them into error classes. These error classes are subjective and based on human intuition -- but despite some ambiguity, we believe that grouping errors is an important step in understanding them, and allows us to analyze them quantitatively.

\begin{figure}[t]
    \centering
    \includegraphics[width=\linewidth]{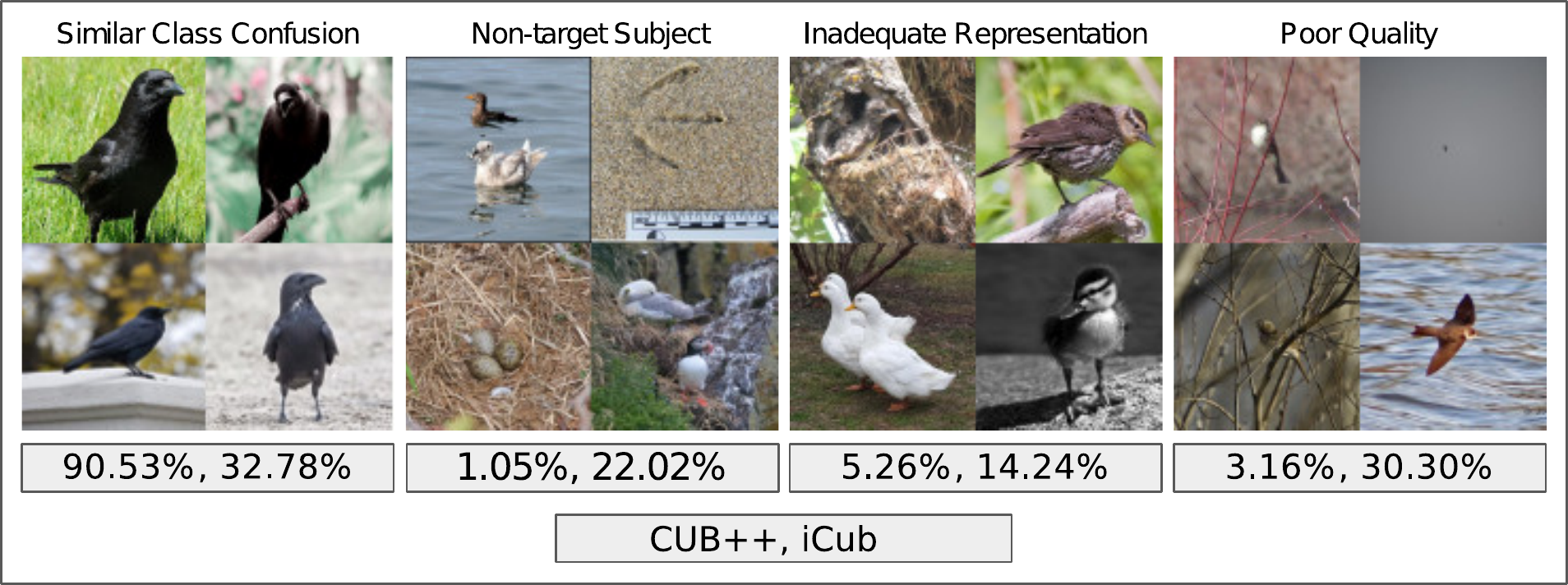}
    \caption{\small\textbf{Examples and Prevalence of Error Groups.} Images from CUB++ and iCub100 help clarify the contents of each error group. The images that no model classified correctly for each dataset were considered, with all such errors for CUB++ and hundreds of such errors for iCub100 being sorted into the 4 groups. Below the images of each group, the percent of errors belonging to each group are shown, first for CUB++, then for iCub100.}
    \label{fig:qual-quant-analysis}
\end{figure}

We present four broad error classes: Similar Class Confusion, Non-target Subject, Inadequate Representation, and Poor Quality. We provide definitions of these classes to help clarify and justify them. Fig.~\ref{fig:qual-quant-analysis} shows several examples from each class, and provides a quantitative evaluation of their prevalence within each dataset.



\vspace{-2.5mm}
\paragraph{\textbf{Similar Class Confusion}} One of the most common causes of classification errors in FGVC is the low level of inter-class variation. In some domains, this variation can be low enough to make classification near impossible. This error class applies to any image that belongs to a class that is difficult to distinguish from anther class, or more often from several other classes. For example, the second hardest class from CUB++ for the SOTA methods in terms of per-class accuracy is the Common tern, at 48.83\%. This bird is very challenging to distinguish from the Arctic, Elegant, and other terns, even for humans.

\vspace{-2.5mm}
\paragraph{\textbf{Non-target Subject}} This class includes errors where the ground-truth subject is not found in the center of the image. This includes images with multiple birds; with birds next to other creatures; where the target bird is off to the side of the image; or where a bird-related object (such as a nest or feather) is the only visible subject. This type of error is significantly more prevalent in real-world data than typical benchmark data, and is more common in iCub100 than in CUB++.

\vspace{-2.5mm}
\paragraph{\textbf{Inadequate Representation}} This class includes errors that are most likely due to appearance factors that are not well represented in the training data. Novel backgrounds and poses both fall into this category. This is sometimes due to flaws in the design of the dataset. With birds, differences in stage of life (i.e. nestling, juvenile, and adult) or gender can correspond to significant differences in both shape and color. While CUB treats species as classes, NABirds solves this to some extent by considering different stages of life and different genders as different classes, when appropriate.

\vspace{-2.5mm}
\paragraph{\textbf{Poor Quality}} Sometimes, images in these datasets are just too difficult to distinguish reliably. This error class includes occlusion, such as when a tree branch blocks a key discriminating feature of a bird from the camera, as well as blurriness, unreasonably poor lighting, and inadequate scale, such as when the bird is far enough away that identifying information is lost.

\vspace{-2.5mm}
\paragraph{\textbf{Other}} For some images, it is more challenging to identify why none of the models could assign the correct label -- though for birds this was rare. We identify these instances where applicable (0.66\% of errors in iCub100 fall under this category, and are not shown in Fig.~\ref{fig:qual-quant-analysis}).

\section{Key Takeaways, Open Problems}

As a result of our analysis, we suggest several important directions for future research.

\vspace{-2.5mm}
\paragraph{\textbf{Expanding domain generalization}} Both CUB and iCub100 contain images from the same set of categories -- 200 different classes of bird. Yet, models trained on the CUB training set can reach accuracies pushing 90\% on the CUB test set, but only around 50\% on iCub100. This suggests that additional effort needs to be put into developing methods that have broader generalization, so that they don't fail when backgrounds, viewpoints, or image composition change slightly.

\vspace{-2.5mm}
\paragraph{\textbf{Correspondence and pose invariance}} A particular challenge to generalization is developing models that understand pose and correspondence. This is a critical piece of the recognition process that doesn't seem to be well captured by strictly statistical learning. Models often fail to generalize to unseen poses, even when it is easy for a human to make inferences based on finding correspondence. This is a key issue that needs to be solved going forward.

\vspace{-2.5mm}
\paragraph{\textbf{Discovering and learning complementary models}} As we have shown, some models learn complementary information which, when combined, can provide a significant boost in performance. A fruitful direction may be to investigate and quantify the degree to which models may complement each other as well as methods for effectively distilling the information. We experimented with performing model distillation\cite{hinton2015distilling} from our ensembles, but the results were underwhelming. Building compact models that capture complementary information is a continuing area of research.

\vspace{-2.5mm}
\paragraph{\textbf{Exploring new data horizons}} Most current work in FGVC focuses on the narrow goal of improving accuracy on a few established benchmark datasets. As we have shown, this view of FGVC doesn't properly account for the wide variety that exists even within a single domain (such as birds). With large-scale, curated image databases such as iNaturalist becoming more prominent, there is more data than ever before for exploring just how good current methods are, and for discovering new avenues for improvement.

\vspace{2mm}

We hope that this work will help inspire others to think carefully about the types of hard problems that still need to be solved in FGVC, and provide some potential directions to begin tackling them.

{\small
\bibliographystyle{splncs04}
\bibliography{FarrellMendeley,addl_refs}
}
\end{document}